\documentclass[a4paper,12pt]{extarticle}
\usepackage{geometry}
\usepackage[T2A]{fontenc}
\usepackage[utf8]{inputenc}
\usepackage[main=english,russian]{babel}
\usepackage{float}
\usepackage{amsmath}
\usepackage{amsthm}
\usepackage{ragged2e}
\usepackage{amssymb}
\usepackage{fancyhdr}
\usepackage{setspace}
\usepackage{graphicx}
\usepackage{colortbl}
\usepackage{tikz}
\usepackage{pgf}
\usepackage{subcaption}
\usepackage{listings}
\usepackage{indentfirst}
\usepackage{tabularx}
\usepackage{makecell}
\usepackage[
backend=biber,
style=numeric,
urldate=long,
maxbibnames=99,
sorting=none
]{biblatex}
\DeclareUnicodeCharacter{202F}{FIX ME!!!!}
\usepackage{array}   % load these first

\usepackage{framed} % For the framed environment
\usepackage{xcolor} % For coloring the box
\usepackage{lipsum}

\usepackage[colorlinks,citecolor=blue,linkcolor=blue,bookmarks=false,hypertexnames=true, urlcolor=blue]{hyperref} 
\usepackage{indentfirst}
\usepackage{mathtools}
\usepackage{booktabs}
\usepackage[flushleft]{threeparttable}
\usepackage{tablefootnote}

\usepackage{chngcntr} % нумерация графиков и таблиц по секциям
\counterwithin{table}{section}
\counterwithin{figure}{section}

\graphicspath{{graphics/}}%путь к рисункам

\addto\captionsrussian{} 

\begin{document}
% ------- Cover (Title) Page -------
\begin{titlepage}
  \centering
  \vspace*{3cm}

  % --- Paper title ---
  {\Huge\bfseries Large Language Models in the Task of \\[0.3em]
              Automatic Validation of Text Classifier Predictions\\[0.3em]
              \par}

  \vspace{2.5cm}

  % --- Author block ---
  {\Large Aleksandr Tsymbalov\par}
  {\Large Mikhail Khovrichev\par}
 \vspace{0.5cm}
  % {\large Department of COMPUTER SCIENCE\par}
  % {\large UNIVERSITY / INSTITUTE\par}
  \vspace{1.5cm}

  % --- Publication target & date ---
  % A manuscript prepared for submission to \textit{JOURNAL NAME}\par
  \vspace{0.5cm}
  \today

  \vfill

  % % --- Optional logo ---
  % \includegraphics[width=0.15\textwidth]{logo.png}\par
  \vspace{0.3cm}

  % --- Corresponding author & contact ---
  \onehalfspacing
  \justifying
  % \textbf{Corresponding author:} YOUR NAME \\
  % Email: \href{mailto:you@example.edu}{you@example.edu}

\end{titlepage}

\section*{Abstract} 
\addcontentsline{toc}{section}{Abstract}  % add Annotation to Table of Contents

Machine learning models for text classification are trained to predict a class for a given text. To do this, training and validation samples must be prepared: a set of texts is collected, and each text is assigned a class. These classes are usually assigned by human annotators with different expertise levels, depending on the specific classification task. Collecting such samples from scratch is labor-intensive because it requires finding specialists and compensating them for their work; moreover, the number of available specialists is limited, and their productivity is constrained by human factors. While it may not be too resource-intensive to collect samples once, the ongoing need to retrain models (especially in incremental learning pipelines \cite{incremental1}) to address data drift (also called model drift \cite{ibm_model_drift}) makes the data collection process crucial and costly over the model's entire lifecycle. This paper proposes several approaches to replace human annotators with Large Language Models (LLMs) to test classifier predictions for correctness, helping ensure model quality and support high-quality incremental learning.

\section*{Keywords}
Large Language Models, LLM, Deep Learning, NLP, Data Annotation, Data Markup, Text Classification, Incremental learning.
\pagebreak

\section{Introduction}
The task of text annotation is popular across numerous domains in which text manipulation must be automated through machine learning. When no off-the-shelf classifier exists for a new dataset, practitioners must first create a gold-standard set of labeled texts to train the model, collect benchmarks to measure quality, and regularly check the classifier for degradation in the metrics (data drift). High-quality data for training, validation, and testing are usually impossible to obtain without specialized human annotators.

With the growing popularity of LLMs, more professionals are now exploring whether humans can be replaced in certain areas of work. Because LLMs possess knowledge and reasoning abilities in many domains and can handle free-format tasks \cite{awesome_prompt1}\cite{awesome_prompt2}, they are strong candidates to augment or even replace human annotation. Simply replacing humans with another text classifier (to validate predictions of the first classifier) would lead to a conventional classification pipeline, requiring frequent retraining of two models.

Human data annotation is used to validate the predictions of the text classifier and to collect high-quality data for further retraining of the classifier. The text classifier predicts a request (intent) in Russian from clients in a company's support chat system. If the classifier cannot correctly classify the intent, it will not initiate the automatic process to close the client request, necessitating a hired specialist to resolve the issue manually. \\\indent
Therefore, it is important to have a high-quality classifier, which in turn requires high-quality annotation. \textbf{This paper focuses on the possibility of replacing human annotation with LLM annotation.}

\section{Related Work}

In research, opportunities for using LLMs in classification tasks can be categorized into two groups: \textbf{Large Language Models as Classifiers (In-Context Learning)\cite{icl}} and \textbf{LLM as Data Annotators}.

\subsection{LLM ICL Classifier Replacement Approaches}
"In-context learning (ICL) classifier replacement" refers to using LLMs to directly perform classification by prompting the model with examples and letting it predicts class labels, instead of training a separate classifier. Recent studies have explored using powerful LLMs in a few-shot or zero-shot manner to replace traditional classifiers. In the paper \cite{breakingbankchatgptfewshot}, the authors compared the LLM with the ICL approach and fine-tuned masked language models on 77-class classification problem (on the banking77 dataset \cite{banking77}). The authors showed that GPT-4 \cite{gpt4}, using a few well-selected examples and a limited number of examples per class, outperformed the considered fine-tuned models, opening the way for building rapid solutions under conditions of limited computational resources and data. One major drawback is the cost of using proprietary models. Although proprietary models become cheaper annually (the authors cited around ~\$1,600 for intensive GPT-4 usage), their use can still be expensive for large companies.

The authors of the paper \cite{yu2023openclosedsmalllanguage} compared LLMs with fine-tuned approaches of classical language models on Named Entity Recognition (NER) and ordinary classification tasks. The authors found that closed-source LLMs (GPT-4 in particular) achieved the highest performance on challenging, "generalization-heavy" tasks, but smaller fine-tuned models often matched or surpassed LLMs on easier datasets. Additionally, pre-trained open-source LLMs (Llama-2 \cite{llama2}) occasionally yielded better results than GPT-3.5 \cite{gpt35}.

In a relevant study \cite{schnabel2025multistagelargelanguagemodel}, the authors explored the potential of replacing humans in the task of classifying query-to-document relationships (matching relevant documents to queries). Their proposed multi-stage approach using GPT-4o \cite{gpt4o} demonstrated that smart prompting and sequential queries with varied prompts could enhance LLM performance without pre-training methods.

\subsection{LLM as Data Annotator}
An alternative line of research uses LLMs to assist or replace humans during data labeling. In these approaches, LLMs generate labels or synthetic training examples subsequently used to train traditional models. LLM annotation can also validate the quality of a classification model by verifying predicted classes.

For instance, authors in \cite{Gilardi_2023} examined replacing human annotators with GPT-3.5 in a zero-shot setting, achieving higher annotation quality than humans in 4 out of 5 task types, increased consistency with carefully collected human annotations, and approximately a 20-fold cost reduction compared to hired annotators. The paper \cite{pangakis2024knowledgedistillationautomatedannotation} utilized LLM-generated annotations (termed distillation) to label training data, confirming that LLM-based labels offer \textit{"a fast, efficient, and cost-effective method"} nearly equaling human-supervised models.

The authors of \cite{ALMOHAMAD20252402} clearly demonstrated that LLM annotation can successfully classify image data (X-rays), resulting in high-quality CNN models. The paper \cite{rouzegar2024enhancingtextclassificationllmdriven} proposed using GPT-3.5 to selectively replace human annotation only when confident, creating confidence thresholds for each dataset. However, the method relies on GPT-3.5 self-estimating its confidence, which can be unreliable \cite{li2024thinktwicetrustingselfdetection}.

\textbf{Common open issues in papers:}
\begin{itemize}
\item Many studies use data that may have already been used in the training samples of most LLMs before the papers were written. For example, datasets used in \cite{breakingbankchatgptfewshot}, \cite{schnabel2025multistagelargelanguagemodel}, and \cite{rouzegar2024enhancingtextclassificationllmdriven} are publicly available and likely part of LLM training. The data in this paper is unique and described in Section~\ref{sec:data}.
\item Low number of classes in classification tasks. Our collected dataset, detailed in Section~\ref{sec:intent_desc}, contains over one hundred unique classes.
\item Most studies use proprietary models (OpenAI), which may be less economical for organizations with large computing resources or those handling sensitive data. The LLMs in this article are publicly available and described in Section~\ref{sec:exp_llm}.
\item Methods typically extract predictions from generated LLM text, making it challenging to isolate classification rejection zones clearly. This paper proposes a probability-based response approach in Section~\ref{sec:prob_approach}.
\end{itemize}

%  --------- --------- --------- --------- --------- BEFORE SENDED TO MODEL 1  --------- --------- --------- --------- --------- ---------

\section{Methodology}\label{Methodology}
% Методология
\subsection{Dataset}\label{sec:data}
\subsubsection{Intent}\label{sec:intent_desc}
\begin{table}[H]
\centering
\caption{Client intent}
\label{tab:example_request}
\begin{tabular}{@{}p{9cm}p{4.5cm}@{}}
\toprule
\textbf{Client request} & \textbf{Intent/Class} \\
\midrule
\textit{How much does product "apple" cost?} & Get product pricing \\ \addlinespace[0.5em] 
\textit{Please provide the cost of product "apple"} & Get product pricing \\ \addlinespace[0.5em] 
\textit{I'm interested in the dollar of "banana"} & Get product pricing \\ \addlinespace[0.5em] 
\textit{"Apple" product characteristics} & Get product details
\\ \addlinespace[0.5em] 
\textit{"Orange" product characteristics} &  Get product details \\
\bottomrule
\end{tabular}
\end{table}

Client intent is the purpose for which they contact support, which can be a question or a request for action from the company. Table~\ref{tab:example_request} provides query examples and their classes. The company identifies approximately 250 unique intents. However, some intents can be similar, complicating initial classification. This issue is addressed through multi-step classifier retraining, distancing contested classes in distribution as retraining proceeds. Nearly every intent has a description, examples fitting the intent, and controversial examples. \\
There are two special intents:
\begin{itemize}
\item Annotator's refusal to classify (\textit{unknown, unk}) — used when classification is impossible.
\item Multiple intent selection (\textit{Confusing, conf}) — used when several answers might fit (not used in actual annotations; see Section~\ref{sec:multiclass_markup}).
\end{itemize}

\subsubsection{Text Classifier Dataset}
To train a classifier from scratch, intent-labelled datasets are periodically collected. it is claimed these datasets maintain high quality through human annotation. They also include daily annotations checking classifier predictions by humans, with dataset sizes typically in the hundreds of thousands.

\subsubsection{Additional Documents}\label{sec:documents}
Some client intents can be resolved using pre-prepared documents, employed by support staff to answer client queries. Document length averages 500 characters, extending up to 2,000 characters if necessary. it is also important to note that the content of some documents may overlap or even conflict each other in some cases. A trained specialist manages these cases effectively.

\subsubsection{Annotation benchmark}\label{sec:bench}
\begin{table}[H]
\centering
\caption{Benchmark Datasets}
\label{tab:bench}
\begin{tabular}{@{}lp{4cm}p{5cm}@{}}
\toprule
\textbf{Dataset} & \textbf{N-rows} & \textbf{Description} \\
\midrule
Binary benchmark & 18,000 & Annotations of binary classifier predictions ("yes"/"no") for text classification correctness \\ \addlinespace[0.5em] 
Multi-class benchmark & 12,000 & Selections from top-5 predicted classes of a text classifier \\
\bottomrule
\end{tabular}
\end{table}

To verify the quality of LLM annotation, quality annotations were collected from \textbf{highly qualified, expensive experts}. Then these data were marked up by the company's regular annotators, who currently annotate the predictions of the text classifier, as well as by LLM annotation. Due to the high cost and lengthy annotation process required by expert annotators, their services cannot be used for regular annotation. The number of unique classes is approximately 250 for benchmarks, with classes distributed according to product distribution (Table~\ref{tab:bench}).\\\indent
The main criterion indicating that LLM annotation performs better than human annotation is: \textbf{all metrics obtained by LLM annotation must be equal to or better than regular human annotations when compared to expert benchmark annotations}. Despite this stringent requirement, the impact of possible annotation errors on classifier performance has also been studied (Section~\ref{sec:res_fpfn}), considering scenarios where solutions might reduce certain classification metrics.

\subsection{Human-grounded annotation}
% 0. Что мы понимаем под человеческой разметкой (как она проводится, что за перекрытия). MK и бинарка
% Что бинарку можно делать сразу после мк, откуда классы для мк
One common use of human annotation is to verify the correctness of predictions made by a text classification model that outputs a large number of classes (a classic text classification scenario). To maintain annotation quality, the same instance is shown to multiple annotators, and an annotation is deemed correct if their answers coincide.\\\indent
Assume that the classification model
can return K best probability predictions, then two types of annotations can be used to evaluate the model's performance.
\subsubsection{Binary Annotation}
Let \(C\) represents the predicted class with the highest probability of the text classifier, \(T\) is client text.
Then:
$$
\mathbf F \colon \mathbf T \times \mathbf C \to \{\text{``0''}, \text{``1''}\}, \quad \text{where } |\mathbf C| \approx 250,\ |\mathbf T| = \infty,
$$
where \(F\) is an annotator who returns a response of 0 (\textit{"no"},\textit{ "the class does not match the text, classification model made an incorrect prediction"}) and a response of 1 (\textit{"yes"}, \textit{"the class match the text, classification model made a correct prediction"}).
\pagebreak
\subsubsection{Multi-class Annotation.}\label{sec:multiclass_markup}
Let $C = \{c_1, c_2, c_3, ..., c_{n}\}$ be the set of all possible classes, and \(T\) be a set of texts.\\
Then:
$$
\mathbf F \colon \mathbf T \times \mathbf C^5 \to \mathbf C, \\
\text{where } |\mathbf C| \approx 250,\ |\mathbf T| = \infty,
$$
\[
\mathbf F\bigl(t,(c_{i_1},\dots,c_{i_5})\bigr)\;\in\;\{c_{i_1},\dots,c_{i_5}\}
\quad\forall\,t\in\mathbf T,\;(c_{i_1},\dots,c_{i_5})\in\mathbf C^5,
\]
where \( C^{5}\) is the set of tuples of length 5 obtained from the classifier as the five most variant classes for the text. \( F\) is an annotator that selects one of these five classes (got from a text classifier) based on the text. \\\indent
Consequently, in multi-class annotation tasks, annotators can resolve ambiguous cases, for example, when the five classes are closely related and the classifier might easily select the wrong one. If the classifier considers that the intents are very similar to each other, the annotator can choose the correct label from the model's top five likely predictions.
Then, as the classifier is further trained on these newly annotated data, the probability margins for ambiguous classes are broadened in general, which follows from the analysis of the cross-entropy loss function used to train the classifier (see Section~\ref{model}). Earlier studies in the company showed that it is enough to take from 1 to 5 predicted probable classes from the classifier for annotators, because this reduces the burden on annotators (also with maximum probability, the true class is within five classes).\\\indent
it is important to note that annotators have another option that may sometimes be appropriate. This is the "request has multiple intentions" class. it is suitable in cases where the client's request cannot be clearly classified, but refusing to classify it is also not appropriate, since the true class is somewhere in the list passed to the annotation (\textit{Confusing}, \textit{"conf"}).
Only human annotators currently have access to this class, and only for analytical purposes and for conducting experiments. In real annotations, they are required to select one of the intents (suggested intent list always includes \textit{"unk"} (\textit{Unknown})).
\pagebreak
\subsection{LLM Annotation}
This paper proposes two ways to obtain the LLM annotation. Depending on your requirements, you can choose one or the another. Combining approaches is not effective; the reasons are described in Section~\ref{sec:prob_after_text}.

\subsubsection{Text-approach}
\begin{figure}[H]
    \centering
    \includegraphics[width=1\linewidth]{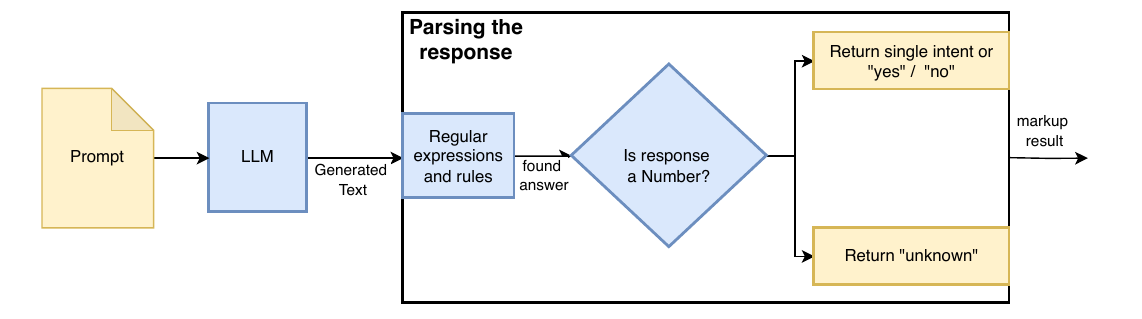}
    \caption{Text-approach}
    \label{fig:text_approach}
\end{figure}
The first approach (Fig.~\ref{fig:text_approach}) is the classic one, based on retrieving the answer from the generated text (text-approach). This approach is described in most of the research because it does not require direct access to the model's probabilities and allows using LLM as a black box. Also, this approach provides an elegant explanation of the LLMs response, which allows modifying the prompt, creating a multi‑stage system with the LLM (e.g., having one LLM check the reasoning of another using different prompts), and enabling the LLM to use tools to improve annotation quality. For example, an LLM can invoke heavy analytical service with additional data only if it is not sure that the current context is sufficient for correct annotating. However, as discussed earlier, this approach makes it difficult to distinguish the refusal‑to‑classify zone. For multi-class annotation, we suggest including the "refuse classification" (\textit{"unk"}) option among the top five predicted classes (note that the classifier itself can also refuse classification; therefore, the top five may already include this option).

\subsubsection{Prob-approach}\label{sec:prob_approach} 
\begin{figure}[H]
    \centering
    \includegraphics[width=1\linewidth]{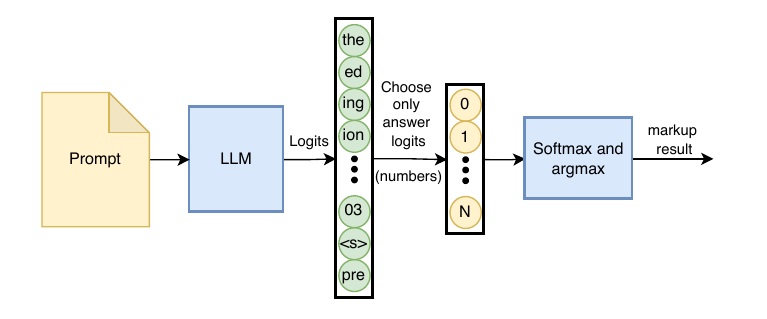}
    \caption{Prob-approach}
    \label{fig:prob_approach}
\end{figure}

The second approach (Fig.~\ref{fig:prob_approach}) is based on obtaining answer probabilities from the LLM (prob-approach). For this purpose, we examine the probability of the first token generated after the prompt. In the case of binary annotation, the LLM is instructed to respond with "0" or "1"; in the case of multi-class annotation, it is instructed to respond with the number corresponding to the specific class listed in the prompt. This probability-based approach not only provides true response probabilities from the LLM at the architectural level, but also enables the introduction of a rejection zone (denoted as \textit{"unk"}) via thresholding; therefore, only examples for which the LLM is confident are annotated. Unfortunately, this approach does not yield model reasonings for option selection, since only a single token is generated. In Section~\ref{sec:prob_after_text}, on the metrics of the text and probability approaches, we show that if reasoning is allowed, the annotation metrics will be lower than when answering with a single token.\\\indent
The thresholding works as follows: among $N$ answers, the maximum probability is taken; if this probability is below the threshold, the classification is rejected (LLM predicts \textit{"unk"}) and sent to human annotators, otherwise the confident answer is returned.

\subsection{Metrics}
\subsubsection{Classifier Metrics}
The classifier is trained to solve the multi-class classification task, that is why all metrics related to this task are relevant to it. For example, F1 metric, accuracy, precision, recall, ROC-AUC, PR-AUC and others. But the two most important metrics are undoubtedly accuracy and coverage. Accuracy refers to the ordinary precision of predictions, and coverage refers specifically to the number of classification refusals (the prediction of a special class that says it is better to transfer a client's request to a hired specialist). Let's call this special class \textit{"unk"} (unknown). The coverage formula is:
   $$
   \text{Coverage} = \frac{1}{N}\sum_{i=1}^{N} \mathbf{1}\{y_{C_i}\neq \text{unk}\},
   $$
where $y_{C_i}$ is classifier prediction.

\subsubsection{Annotation metrics}

In the annotation task, the only effective way to assess an LLM annotation capability is to measure how often its predictions coincide with those of human annotators, since the models initially have access only to human‐provided labels. Confidence in human annotation can be increased by collecting labels from several annotators on the same instance and retaining it only if they all agree, thus forming a high‐quality test set.
Let $N$ be the total number of samples, $y_i$ be the prediction of annotator ${H}$ (human) or ${M}$ (LLM). Since human annotation on the benchmark was collected only twice (data collection from experts and data collection from regular annotators), we will fix $H$. \\The accuracy of the LLM annotator $M$ is:
% \[
% \text{Accuracy}(M, H)
% = \frac{1}{N}\sum_{i=1}^{N} \mathbf{1}(y_{M_i} = y_{H_i}),
% \]
  $$
  \operatorname{Accuracy}_H(M)\;=\;\frac1N\sum_{i=1}^{N}\mathbf 1\!\bigl(y^{(M)}_i = y^{(H)}_i\bigr),
  $$
where \(\mathbf{1}(\cdot)\) is an indicator function that equals 1 if \(y_{M_i} = y_{H_i}\), and 0 otherwise.

it is also meaningful to measure the fraction of "abstentions," i.e., cases in which the LLM annotation is "unknown" (\textit{"unk"})
Suppose there is \(N\) the total number of predictions by model \(M\), and suppose there exists a class labeled "unknown" (\textit{"unk"}) Denoting the model's predictions by \(y_{M_i}\):
\[
\text{Coverage}(M)
= \frac{1}{N}\sum_{i=1}^{N} \mathbf{1}(y_{M_i} \neq \text{unk}),
\]
where \(\mathbf{1}(\cdot)\) is 1 if \(y_{M_i} \neq \text{unk}\) and 0 otherwise. Since human annotators may also refuse to label certain instances, human coverage can likewise be compared with LLM coverage. \\
Finally, we can combine accuracy and coverage into a single overall metric:
\[
\text{Accuracy}_{\mathrm{total}}(M)
= \text{Accuracy}(M)\times \text{Coverage}(M),
\]
which measures the fraction of instances that are both non‐abstained and correctly labeled.\\
\textbf{We can also treat the \textit{"conf"} class ("multiple intentions") as equivalent to \textit{"unk"} for LLMs metric calculation, since it appears only in experimental human annotations}. In binary annotation tasks, the usual F1 score (macro avg., because both classes are equally important), precision, and recall remain appropriate. \\
\textbf{Note that:}
\begin{enumerate}
    \item Coverage and accuracy are calculated in the same way independent of \textit{сonf} for the LLM.
\item \textit{"unk"} in annotation and \textit{"unk"} as a classifier prediction are not full identical: the classifier learns to predict \textit{"unk"}, while an annotator chooses \textit{"unk"} only when no other label fits.
\end{enumerate}

\subsection{Text classification model}\label{model}
\subsubsection{Architecture}
A text model refers to a model consisting of encoders with a trainable full-link output layer. Examples: BERT type models. A text classifier can actually be any mathematical model.
In our proposed approach, \textbf{the classification model is not only fine-tuned on annotated texts but is also used as a re-ranker}, returning only a limited subset of classes from which annotators select the correct label. it is also worth considering that the classification model cannot actually be a reranker in the usual definition, because it is trained at predicting exactly the true class and not to rank all other classes by relevance. 
% \subsubsection{Impact of markup on the classifier} про импакт было сказано в разделе с текстовым датасетом

\subsubsection{Loss}\label{sec:loss}

The text classification model is trained to predict one of $N$ classes using the \textbf{cross-entropy loss}
$$\mathcal{L}_{\text{CE}} = -\sum_{i=1}^{C} p_i \log\left(\frac{\exp(z_i)}{\sum_{j=1}^C \exp(z_j)}\right),$$
where \(z_i\) are the logits for each class, and \(p_i\) is the class label. Model classifier is used to classify the intention of clients. it is obvious that some number of training samples may contain errors (due to complexity of samples or human factor). Replacing human annotation with LLM annotation may result in a change in the quality of the trained model that is why further in the paper there are experiments to measure the effect of different kinds of errors on the classifier (Section~\ref{sec:res_fpfn}).
\pagebreak
\subsection{Prompt}
\noindent Prompts are presented in Section~\ref{sec:appendix_prompt}.
\subsubsection{Zero-Shot and Few-Shot}
LLM is capable of solving problems without any examples \cite{kojima2023largelanguagemodelszeroshot}, only based on the knowledge learned during training. But as practice of application and many researches show, if you give some examples of problem solving \cite{fewshot}, the quality of solved problems and following the instructions increases. Positive examples are defined for each intent, negative examples are generated specifically for this work as examples that are most difficult to classify by models and humans.
% Further experiments are presented where it is tried to provide not examples of already performed markings, but examples of texts relevant for intents, adding to the prompt not only a description of the intent.

\subsubsection{Prompt description}
\subsubsection*{Text-approach prompt (Reasonings)}\label{sec:text_approach_prompt}
In order for the LLM to follow the instructions a prompt was written consisting of the following blocks:
\begin{itemize}
    \item \textbf{Domain introduction.} Some information is given about the company, the clients, why it is important to do accurate annotation and check answers.
    \item \textbf{Data details.} The concept of intent is explained, as well as the fact that they come from a text classifier.
    \item \textbf{Task.} Depending on the modifications, the task includes at least the client's text and a description of the intent that needs to be checked to see if it fits the class, or a description of several intents from which only one suitable description needs to be selected. Next, relevant documents can be found (Section~\ref{sec:rag}) for the client's text, which may contain the answer to their question, explanations of terminology, and other useful information, as well as high-quality examples (positive and negative) of texts for intents.
    \item \textbf{Input and output format.} The LLM receives the task in the \textbf{[USER]} block. Then all reasoning should be done in the \textbf{[REASONING]} block and the final answer should be given in the \textbf{[ANSWER]} block. In case of binary annotation task, the final answer is 'yes' or 'no', in case of multi-class annotation the final answer is one selected intent from the list passed in the task.
    \item \textbf{Reasoning.} The prompt mentions that reasoning should go step by step, reveal the meaning of unfamiliar words in the client's request, use all the information provided, but not imagine it for the client.

\end{itemize}

\subsubsection*{Prob-approach prompt}
The difference from the text-based prompt is that the prob-approach does not require an emphasis on reasoning. Also, after the instruction and task are given, before getting the probabilities of answers, the \textbf{[ANSWER]} block is added to the LLM response block (chat format depends on the specific LLM); therefore, it is enough for LLM to generate only 1 character, which can be either 0 or 1 (if binary annotation), or a number from 1 to 5 (in the case of multi-class).

% \subsubsection{Critic-Prompt}
% The LLM has a problem with agreeing with statements \cite{agree}. This is especially true in the case of binary annotation, where there are only two answers "yes" and "no"; therefore, if the intent to be checked is at least very similar to the client's text, there is a high chance of getting an error of the first type (False Positive). In an attempt to reduce this kind of error, a prompt was written to force more double-checking of oneself \cite{selfrefine} and less trust in examples and sources. 
% It will be shown next that with this approach it was possible to increase the number of correctly labeled labels of class "no", but not more than with the usual addition of negative examples

\subsection{Alignment (SFT)}
Two approaches were used to improve annotation quality: soft fine‑tuning on the classification task (adding a classification layer directly to the LLM) and soft fine‑tuning to enhance the LLM reasoning ability for correct annotation. To avoid modifying all LLM parameters due to computational resource constraints, we employ a low‑rank adaptation (LoRA) \cite{lora}
$$
\;\widetilde{\mathbf W} \;=\; \mathbf W_0 \;+\; \Delta\mathbf W\;, \!\
\Delta\mathbf W \;=\; \frac{\alpha}{r}\;\mathbf{B}\mathbf{A},
$$
\[
\mathbf W_0\in\mathbb R^{d_{\text{out}}\times d_{\text{in}}},\quad
\mathbf A\in\mathbb R^{r\times d_{\text{in}}},\quad
\mathbf B\in\mathbb R^{d_{\text{out}}\times r},
\]
where $\widetilde{\mathbf W}$ is the new weights achieved, $\mathbf W_0$ is the frozen weights, $\alpha$ is a scalar LoRA and $r$ is a rank.
\subsubsection{SFT classification}\label{sec:sft_classification}
To further improve annotation metrics, we fine‑tune the LLM directly on the annotation task. First, annotation data are collected; then the LLM is modified to function as a classifier by adding a linear layer (classification head). The hidden state of the LLMs last token is fed into this head, after which the resulting LLM–classifier is trained like a standard classifier (see Section~\ref{sec:loss} for the loss function). All original LLM weights remain frozen except for the LoRA parameters and the classification head weights. This approach restricts the LLMs reasoning abilities but yields stable, high‑quality annotation metrics, which is advantageous for automation.

\begin{table}[H]
\centering
\caption{Data for SFT (classification)}
\label{tab:datasets_sft_classification}
\begin{tabular}{@{}lp{4cm}p{5cm}@{}}
\toprule
\textbf{Dataset} & \textbf{N-rows} & \textbf{Description} \\
\midrule
Multi-class training data & 50,000 & Training data with approximately equal distribution of the number of intents \\ \addlinespace[0.5em] 
Multi-class validation data & 20,000 & Validation data with prod distribution of intents \\
\bottomrule
\end{tabular}
\end{table}
As can be seen in Table~\ref{tab:datasets_sft_classification}, training and validation data were collected from the production data stream, but in the training sample, the number of classes was balanced.

\subsubsection{SFT reasonings}\label{sec:sft_reasoning_method}
% Get gpt-4o and R1 reasoning, tune 
In order to improve the quality of LLM reasoning (which is unable to reason qualitatively in its basic implementation) high-quality reasoning that has been correctly labeled can be collected from humans or more complex and closed LLMs, after which additional fine-tune can be performed. Such soft fine-tune can not only improve the quality of annotation when using a text-based approach, but also increase the quality of annotation when using a prob-approach, since the final representations of the initial instruction and task after training will also differ, probably towards the best representation of information, even without further reasoning generation from the LLM. 
A hypothesis that pretraining in general is also improve the annotation metrics for the prob-approach proved to be true, as demonstrated further in Section~\ref{sec:sft_reasonings}.
\begin{table}[H]
\centering
\caption{Data for SFT (reasonings)}
\label{tab:datasets_sft_reasonings}
\begin{tabular}{@{}lp{4cm}p{5cm}@{}}
\toprule
\textbf{Dataset} & \textbf{N-rows} & \textbf{Description} \\
\midrule
Binary training data & 8,000 & Training data with approximately equal distribution of the number of intents \\ \addlinespace[0.5em] 
Binary validation data & 2,000 & Validation data with prod distribution of intents \\
\bottomrule
\end{tabular}
\end{table}
As we can see in Table~\ref{tab:datasets_sft_reasonings}, there is significantly less data with reasoning than data for SFT (classification). This is because collecting reasoning is much more expensive than obtaining high-quality labels in multi-class or binary task format.

\subsection{RAG}\label{sec:rag}
\begin{figure}[H]
    \centering
    \includegraphics[width=1\linewidth]{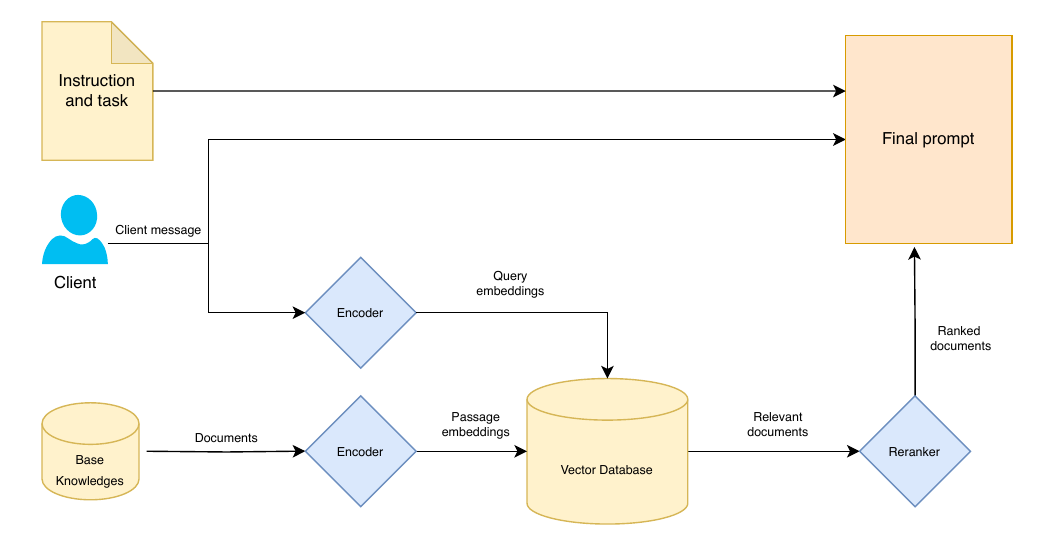}
    \caption{RAG example}
    \label{fig:rag_example}
\end{figure}

Not all company-specific terms may be known to LLM, as well as some client questions that can actually be answered using methodological materials written for support chat operators. To increase LLMs awareness when solving a task, a RAG system (Fig.~\ref{fig:rag_example}) was developed to search for retrieve documents relevant to the client’s request.  

\subsubsection{Design}

RAG typically comprises two components: retrieval and re‑ranking. Retrieval selects a set of documents based on simple metrics (e.g., cosine similarity between encoder embeddings of the query and the documents). Although retrieval is fast, it does not order documents precisely by relevance. Retrieval can also be performed using statistical algorithms without neural networks (e.g., BM25 \cite{bm25}). Retrieved documents are usually stored in a vector database. These documents can then be re‑ranked by a more complex model (cross‑encoders) which helps to select the most relevant documents and order them by importance to the query. Large documents are split into smaller ones using various techniques (by paragraph, special symbols, or sentence-similarity within a chunk). This chunking ensures each piece is small enough to include in full without further division.

\subsubsection{Prompt Design}
Relevant documents are placed in the prompt, after which the LLM solves the task in the usual format.
The prompt instruction changes slightly, with the addition of the \textbf{[RETRIEVED]} block, explaining to the model that this block will contain information that may help to answer the client's query. In the task-block, after the description of the intents and examples, documents are added below the \textbf{[RETRIEVED]} block.

\subsection{Proposed Approach}
\begin{figure}[H]
    \centering
    \includegraphics[width=1\linewidth]{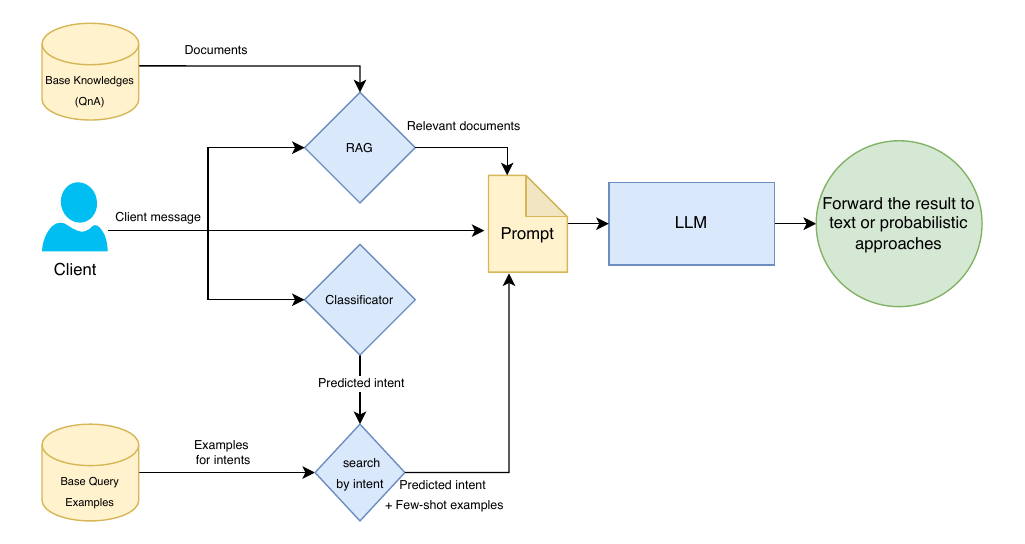}
    \caption{Proposed pipeline}
    \label{fig:pipeline}
\end{figure}
\noindent
The final proposed approach is shown in Fig.~\ref{fig:pipeline}:
\begin{itemize}
    \item First, the client's request is sent to the classifier and RAG.
    \item The RAG system searches the knowledge base for relevant documents, which are then sent to the prompt.
    \item The classifier model predicts a class to the client request, then sends one or more predicted intents to a dictionary where the key is the intent and the values are examples of the intent and description. The values are then forwarded to the prompt
    \item The final prompt is formed from the instructions, relevant documents, and information about the intents. The prompt is sent to the LLM.
    \item Depending on the approach, we take either the probabilities of answers from the LLM (probabilistic-approach) or the explanation and the answer itself (text-approach).
\end{itemize}
Ensembles can be used to modify this approach. This requires obtaining token probabilities using a prob-approach for multiple LLMs.

\subsection{Ensembles}
\begin{figure}[H]
    \centering
    \includegraphics[width=1\linewidth]{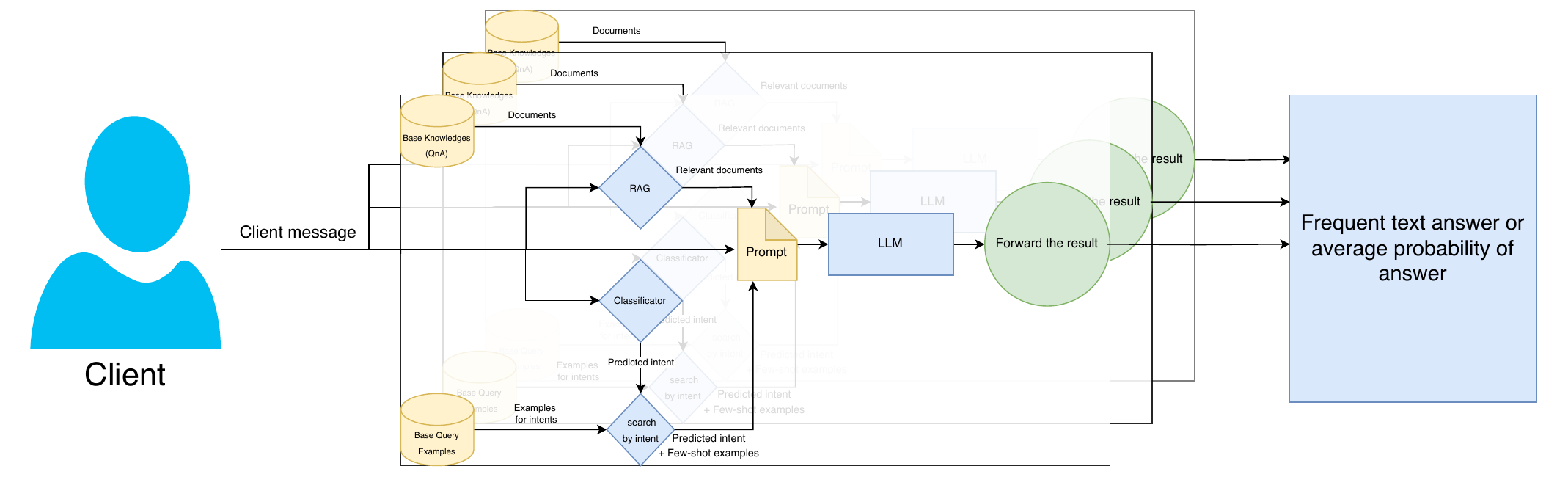}
    \caption{Simple ensemble}
    \label{fig:ensemble_example}
\end{figure}
A simple example of an ensemble is shown in Fig.~\ref{fig:ensemble_example}. Obviously, it is not necessary to run RAG multiple times, nor is it necessary to run a search for examples for intents. However, if there are multiple pipelines with different RAG configurations and different types of examples for intents, then the scheme in the figure is relevant.

\subsection{Annotation completion time}
Previous experiments conducted by the company showed that 1 multi-class annotation takes \textbf{13.4} seconds on average, while 1 binary annotation takes \textbf{6.8} seconds on average (these statistics take into account that several annotators can annotate tasks in parallel). When using vLLM \cite{vllm} and 1 GPU per 1 multi-class, the best approach with the ensemble takes \textbf{~0.85} seconds, which is \textbf{8 times faster} than the binary annotation task and \textbf{15.7 times faster} than multi-class annotation. \\\indent
For example: multi-class benchmark took a human annotator almost two days to annotate, while LLM annotation took only 3 hours. The problem of annotation duration is particularly acute when several parallel annotations are running in a company. The ability to run LLM annotation on multiple GPUs \textbf{reduces the queue and speeds up the entire annotation process.}

\section{Methodological Notes}
\subsection{Parameters}
All parameters were searched through on separate samples without test data leakage or adaptation to the test sample. To select the parameters, the load data was collected from production and annotated separately; the benchmark was used exclusively for final comparisons.
\subsection{Completeness of experiments}
If algorithms in one type of annotation (binary or multi-class) show a clear advantage in experiments, they are not repeated for another type of annotation, with some exceptions. The reason for this is to reduce the amount of computation.
\subsection{Threshold search policy for a prob-approach}
The threshold values for the prob-approach were also chosen therefore the approach would show results higher than human annotators on annotation metrics, but not the highest possible, since rejecting classification logically reduces coverage and increases precision (because only examples that the LLM is confident about are annotated).
Since people also tend to make mistakes, if the annotation metrics are higher for LLM, the classification metrics will not only be closer to the true values when periodically measuring the quality of the classifier, but the trained classifier will also perform better on new data (Section~\ref{sec:res_fpfn}). \\\indent
It may be possible to achieve even better results by combining approaches that were not described in detail in the experiments, or by carefully tuning all hyperparameters, but the main goal of the experiments was achieved: the possibility of replacing humans with LLM without loss of quality in the annotation task was demonstrated.

\pagebreak
\section{Experiment Setup}
\subsection{LLMs}\label{sec:exp_llm}
\begin{itemize}
    \item Llama3.3-70b \cite{llama} \\
    Dense decoder-only Transformer (RMSNorm+SwiGLU) with 128k-token vocabulary, pre-trained on 15T multilingual tokens and post-trained with SFT + Rejection Sampling + Direct Preference Optimisation. \\
    Native capabilities: multilingual Q\&A, coding, reasoning and tool-use hooks.
    \item Qwen2.5-32b \cite{qwen} \\
    32b dense Transformer with revamped tokenizer and 18T-token pre-training corpus \\
    Model is purely dense, making it drop-in for GPU inference.
Training focus: cold-start data + reinforcement learning that explicitly rewards chain-of-thought quality, bringing reasoning scores close to OpenAI-o1.
    \item Vikhr NeMo-12b \cite{vikhr} \\
    Bilingual Russian-English instruction Mistral-Nemo\cite{mistral} with a 40k SentencePiece vocab adapted to Russian. \\
Pipeline: rebuild tokenizer, 11b-token continued pre-training with KL regularisation to avoid catastrophic forgetting,  instruction tuning.
    \item Gemma2-27b \cite{gemma} \\ Decoder-only Transformer with Grouped-Query Attention and interleaved local-global attention, GeGLU activations and RMSNorm throughout.
    \item Athene-V2-Chat-72b \cite{athene} \\
    Fine-tuned from Qwen2.5 72b with a "targeted post-training" RLHF pipeline; released under open weights by Nexusflow. \\
    Specialties: strong function-calling agent variant and long-log extraction accuracy, aimed at enterprise tool-use scenarios. 
\end{itemize}
LLM can be used with a text-based approach or a prob-approach, with or without RAG, and can be fine-tuned using LoRA. There are 4хA100s available.
\subsection{RAG}
A combination of retrieval and reranking, selection of the number of documents to add to the prompt, combination with a prob-approach were conducted. Main parameters reviewed: number of documents appearing in the prompt, document and query similarity thresholds, but only the number of documents is of particular value.\\
\textbf{Experiments}: 
\begin{itemize}
\item Selecting documents relevant to the client's query sent directly to the chat (Table~\ref{tab:rag_multiclass}).
\item Selecting documents relevant to the verifiable intent or intents in the annotation (Table~\ref{tab:rag_multiclass_intent}).
\end{itemize}
\subsubsection{Retrievals}
\begin{itemize}
    \item multilingual-e5-large-instruct \cite{e5}\\
    Instruction-Tuned Multilingual Embedder supports ~100 languages, built on XLM-RoBERTa-large. it is fine-tuned with an instruction format. 
    \item BM25 \cite{bm25}\\ 
    Okapi BM25 is a classic ranking function for document retrieval based on term matching. It scores documents by term frequency in the document and inverse document frequency of query terms, with length normalization. This yields a bag-of-words relevance score for a given query. 
\end{itemize}
\subsubsection{Rerankers}
    \begin{itemize}
        \item bge-m3 \cite{bge_m3} \\
        Multi-Function Multilingual Embeddings is a Beijing Academy of AI model emphasizing Multi-Functionality, Multi-Linguality, and Multi-Granularity. The model can produce BM25-like term weights alongside embeddings, enabling hybrid search with no extra cost.
        \item FRIDA \cite{frida} \\
        FRIDA is a general-purpose text embedding model inspired by a T5 denoising encoder. 
    \end{itemize}
\subsubsection{Embedding storage}
Document embeddings are stored in the faiss IndexFlatIP \cite{faiss}. This index stores all vectors in a "flat" form, without additional structure or compression, and when searching, simply calculates the scalar product between the query and each vector in the index. This approach guarantees accurate search for the closest neighbors according to the scalar product metric.

\subsection{Alignment Experiments (SFT)}
For SFT, LoRA with Llama3.3-70b (for classification tasks) and Qwen2.5-32b (for reasoning soft fine-tuning) were used. The main parameters (rank, alpha) were reviewed for LoRA. Chosen attention projections: $\mathbf{q_{proj}}$, $\mathbf{k_{proj}}$, $\mathbf{v_{proj}}$, $\mathbf{o_{proj}}$
\subsubsection{SFT (classification)}
For further training of the LLM on the classification task from the product stream, data was described and modifications to the LLM were made in Section~\ref{sec:sft_classification}. After training, the results were evaluated on multi-class benchmark from Section~\ref{sec:bench}. 

\subsubsection{SFT (reasonings)}
The collected training data using GPT-4o \cite{gpt4o} and Deepseek-R1\cite{deepseek} were used to SFT the Qwen2.5-32b model. The reasoning of other Large Language Models was parsed further, with the reasoning placed in the \textbf{<think></think>} block and the answer placed in \textbf{<answer></answer>}. Generations were only used for training if they led to the correct answer (the LLM answer matched the answer of professional annotators with extensive experience). it is also important not only to be able to reason well about the question, but also to be able to work with the information given in the task. \textbf{Therefore, documents are also submitted in the prompt using RAG} (Section~\ref{sec:rag}) in the \textbf{<retrieved></retrieved>} block. It is important to note that the training data does not overlap with the benchmark data.

\subsection{The impact of FN and FP on the classifier}\label{sec:setup_fnfp}
To verify the impact of different types of errors in the dataset on the quality of the classifier trained on it, two experiments were designed:
\begin{itemize}
    \item To evaluate the impact of False Positive errors, the class in a certain percentage of records in the training sample was changed to the closest in meaning, but not random. For this process, simple classifier error statistics and manual partitioning were taken.
    \item To evaluate the impact of False Negative errors, a certain percentage of records in the training sample were deleted entirely or changed to the "classification rejected" (equated to \textit{"unk"}) class (in fact, only a certain proportion of this class is present in the sample; therefore, some of the records are truncated).
\end{itemize}

\section{Results}
\subsection{The impact of False Negative and False Positive on the classifier}\label{sec:res_fpfn}

\subsubsection{False Positive}
\begin{figure}[H]
    \centering
    \includegraphics[width=0.7\linewidth]{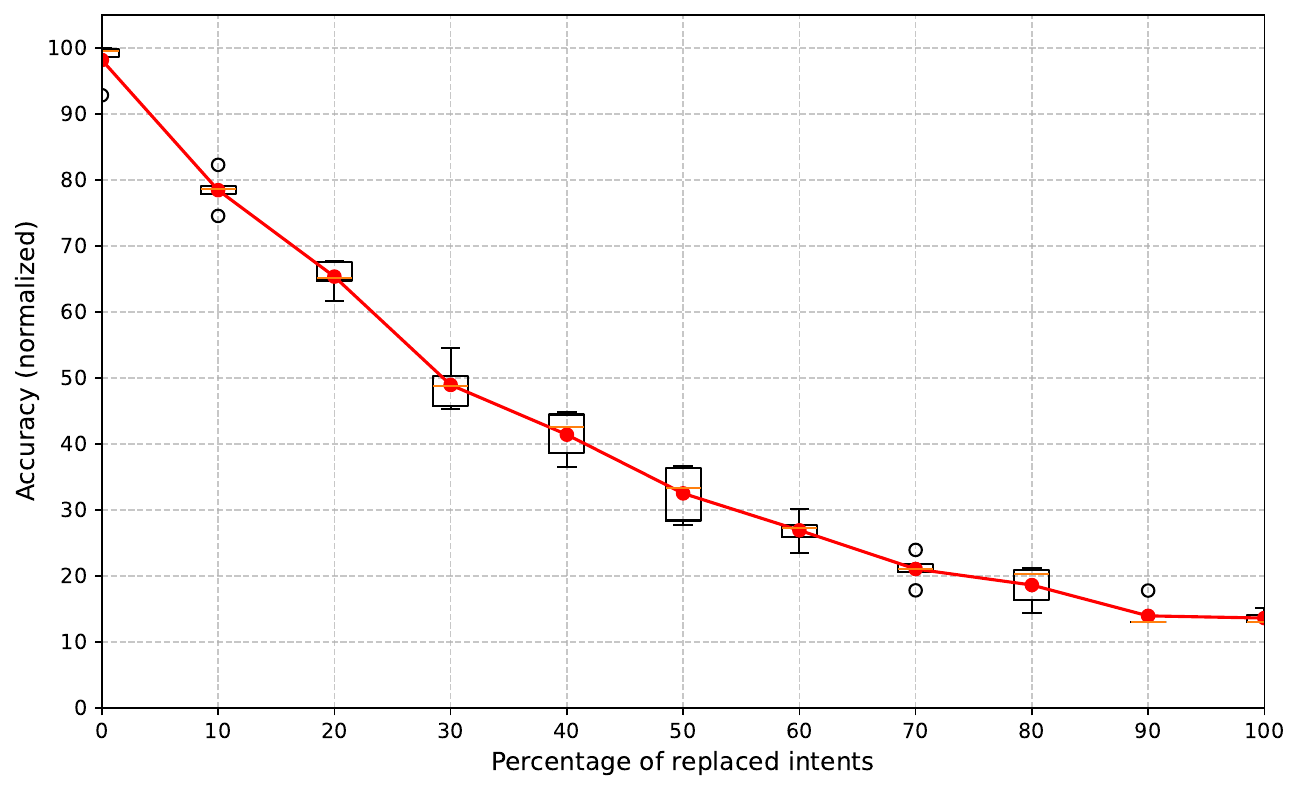}
    \caption{Impact of False Positive on the classifier (5 experiments)}
    \label{fig:fp}

\end{figure}

If we replace the intent in the training sample with the closest similar one and train the text classifier on this, the quality of the test sample begins to deteriorate in a predictable manner. Starting at 70\% replacements (Fig.~\ref{fig:fp}), the quality begins to drop rapidly, but does not reach zero. This may indicate several things:
\begin{itemize}
    \item One text in the data may correspond to several intents; therefore, the closest replacements do not significantly impair the classifier's training.
    \item The marked data contains errors made by the annotators, and replacing it with the closest intent actually corrects these errors, there are enough corrected examples in the dataset for normal training.
\end{itemize}

\subsubsection{False Negative}
\begin{figure}[H]
    \centering
    \includegraphics[width=0.7\linewidth]{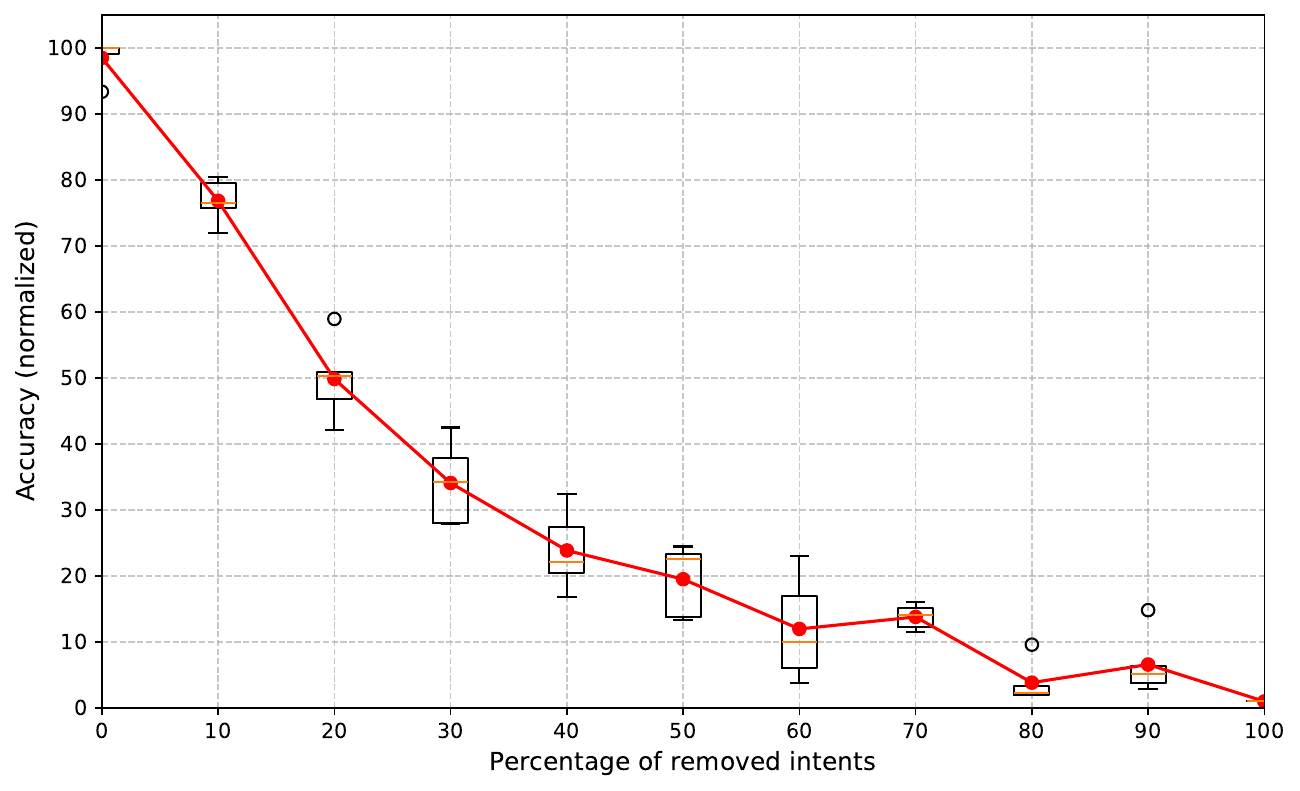}
    \caption{Impact of False Negative on the classifier (5 experiments)}
    \label{fig:fn}
\end{figure}
Removed examples (Fig.~\ref{fig:fn}) in the annotation have a stronger impact on the accuracy of the classifier; accuracy drops faster when the dataset is affected than in the False Positive example. \\\indent
it is likely that the lack of removed examples in intents with a small number of classes greatly affects the model's ability to learn them, which is why the problems become more noticeable in the test sample. it is obvious that with zero records in the training sample, no quality can be achieved. An experiment was also conducted to increase the proportion of \textit{"unk"} in the training dataset, but this only resulted in the classifier starting with some percentage of replacements predicting exclusively unk.

\subsubsection{The importance of mistakes}
After a series of experiments, it became clear that it was necessary to achieve a quality that was not inferior to human annotators across all selected metrics, since the classifier is affected by even the first observable errors in the dataset, and the impact on business metrics requires more in-depth analysis. If LLM can annotate data more accurately than human annotators (i.e., correctly select 1 out of 5 intents or check a specific intent for accuracy), then we can expect to improve the quality of the classifier itself when training on higher-quality data. Also, it is worth considering that reducing the number of False Negative has a more positive effect on the classifier.

\subsection{Vanilla comparison of LLMs}
\subsubsection{Multi-class}\label{sec:vanila_multiclass}
\begin{table}[H]
  \centering
  \footnotesize
  \caption{Metrics for LLMs (Multi-class)}
  \label{tab:vanila_multiclass}
  \begin{tabularx}{\textwidth}{@{}l *{5}{c}@{}}
    \toprule
    \textbf{Model} &
    \makecell{\textbf{Coverage}\\(\textit{conf}=\textit{"unk"})} &
    \makecell{\textbf{Coverage}\\(\textit{conf}\,\(\neq\)\,\textit{"unk"})} &
    \makecell{\textbf{Accuracy}\\(\textit{w/o conf})} &
    \makecell{\textbf{Accuracy}\\(\textit{conf}=\textit{wrong})} &
    \textbf{$\text{Accuracy}_{\mathrm{total}}$} \\
    \midrule
    \textbf{Human}    & 0.2134 & 0.5240 & \textbf{0.6977} & 0.3113 & 0.1489 \\

    \midrule
    \makecell[l]{Vikhr NeMo-12b\\text-approach}      & 0.6254 & 0.6254 & 0.4250 & 0.4250 & 0.2658 \\   
   \midrule
    \makecell[l]{Gemma2-27b\\text-approach}    & 0.4688 & 0.4688 & 0.4931 & 0.4931 & 0.2312 \\
    \midrule
    \makecell[l]{Athene-V2-Chat-72b\\text-approach}    & 0.5498 & 0.5498 & 0.4001 & 0.4001 & 0.2200 \\
    \midrule
    \makecell[l]{Llama3.3-70b\\text-approach}      
     & \textbf{0.6585} & \textbf{0.6585} & 0.4118 & 0.4118 & 0.2712 \\
    \midrule
    \makecell[l]{Qwen2.5-32b\\text-approach}                & 0.5243 & 0.5243 &  0.5399 &  \textbf{0.5399} & \textbf{0.2831} \\
    \bottomrule
    
  \end{tabularx}
\end{table}

The results of the benchmark measurements are shown in Table~\ref{tab:vanila_multiclass}. During the experiments, it was decided to focus further work on two models, as they showed high and stable metrics, good instruction following, and an interesting contrast between coverage and accuracy: Llama3.3-70b and Qwen2.5-32b. Llama3.3-70b shows high coverage with lower accuracy, indicating that it avoids \textit{"unk"} more often and is less uncertain, unlike Qwen2.5-32b, which selects intents more often when confident and returns unk when uncertain.\\\indent
From the beginning of the experiments, it is clear that human annotators are more accurate in multi-class annotation tasks than LLM.

\subsubsection{Binary}
\begin{table}[H]
  \centering
  \footnotesize
  \caption{Metrics for LLMs (Binary)}
  \label{tab:vanila_binary}
  \begin{tabularx}{\textwidth}{@{}l *{7}{c}@{}}
    \toprule
    \textbf{Model} &
    \makecell{\textbf{Prec.}\\\textit{1, pos.}} &
    \makecell{\textbf{Prec.}\\\textit{0, neg.}} &
    \makecell{\textbf{Rec.}\\\textit{1, pos.}} &
    \makecell{\textbf{Rec.}\\\textit{0, neg.}} &
    \textbf{Acc.} &
    \makecell{\textbf{f1-score}\\ (macro avg.)} &
    \makecell{\textbf{Coverage}\\ (perc.)} \\
    \midrule
    \textbf{Human}    & 0.8748 & 0.7162 & 0.7834 & 0.8297 & 0.8018 & 0.7977 & 100\% \\
    \midrule
    \makecell[l]{Llama3.3-70b\\text-approach} 
    &0.8779 & \textbf{0.8178} &\textbf{ 0.9743} & 0.4600 & \textbf{0.8711} & 0.7562&100\%\\
\midrule
    \makecell[l]{Llama3.3-70b\\prob-approach\\without thresholds}               & 0.8718 & 0.8067 &  0.9721 &  0.4496 & 0.8643 & 0.7483 & 100\%\\
    \midrule
    \makecell[l]{Llama3.3-70b\\prob-approach\\use thresholds}                & 0.8918 & 0.6825 & 0.7387 & \textbf{0.8624} & 0.7875 & 0.7850 & 57\% \\
    \midrule
    \makecell[l]{Qwen2.5-32b\\text-approach}                & 0.8613 & 0.6531 & 0.8726 & 0.6307 & 0.8059 &  0.7543 & 100\% \\
    \midrule
    \makecell[l]{Qwen2.5-32b \\prob-approach\\without thresholds} & 0.8471 &    0.6365 &  0.8837 & 0.5606 & 0.7977 & 0.7306     & 100\%      \\
    \midrule
    \makecell[l]{Qwen2.5-32b \\prob-approach\\use thresholds}                   &\textbf{ 0.9398 }& 0.6863 & 0.8518 & 0.8560 &  0.8530 & \textbf{ 0.8277} & 54\%\\
    \bottomrule
  \end{tabularx}
\end{table}
Two LLMs selected from complex multi-class annotation (Section~\ref{sec:vanila_multiclass}) were used in binary annotation (Table~\ref{tab:vanila_binary}). Initially, Llama3.3-70b showed an advantage in the text-based approach over Qwen2.5-32b in almost all metrics. If we try the prob-approach without threshold, we can see a decrease in all metrics for both models. This is because the prob-approach without thresholds disables the LLM's ability to reason, forcing the LLM to write the answer immediately. If we choose the thresholds on a separate dataset to LLM for certainty of answer, we can see the increasing in annotation metrics. This is due to the refusal to annotate examples in which the LLM cannot give a confident answer.\\\indent
With the chosen threshold, LLMs approach human annotators in many metrics. There is a noticeable problem on the negative class, where Llama3.3-70b has low negative precision and recall, and Qwen2.5-32b has low negative precision. At this stage it is not possible to replace a human annotator and it is not profitable to lower the coverage.

\subsection{Prob approach after reasonings}
For consistency, let's use binary annotation, where LLM can choose option 0 ("no") and option 1 ("yes"). If we try, after the text-based approach, to remove the final LLM answer after the [ANSWER] block and leave only the reasoning on the topic, taking only those arguments that do not mention the answer directly but list all possible and improbable answering options, and then run the prob-approach, we will obtain a prob-approach after reasoning. Since the main point of using a prob-approach is the convenience of choosing a classification threshold, it is important to look at how the distribution of answer probabilities changes in the new approach.
\label{sec:prob_after_text}
\begin{figure}[H]
    \centering
    \includegraphics[width=1\linewidth]{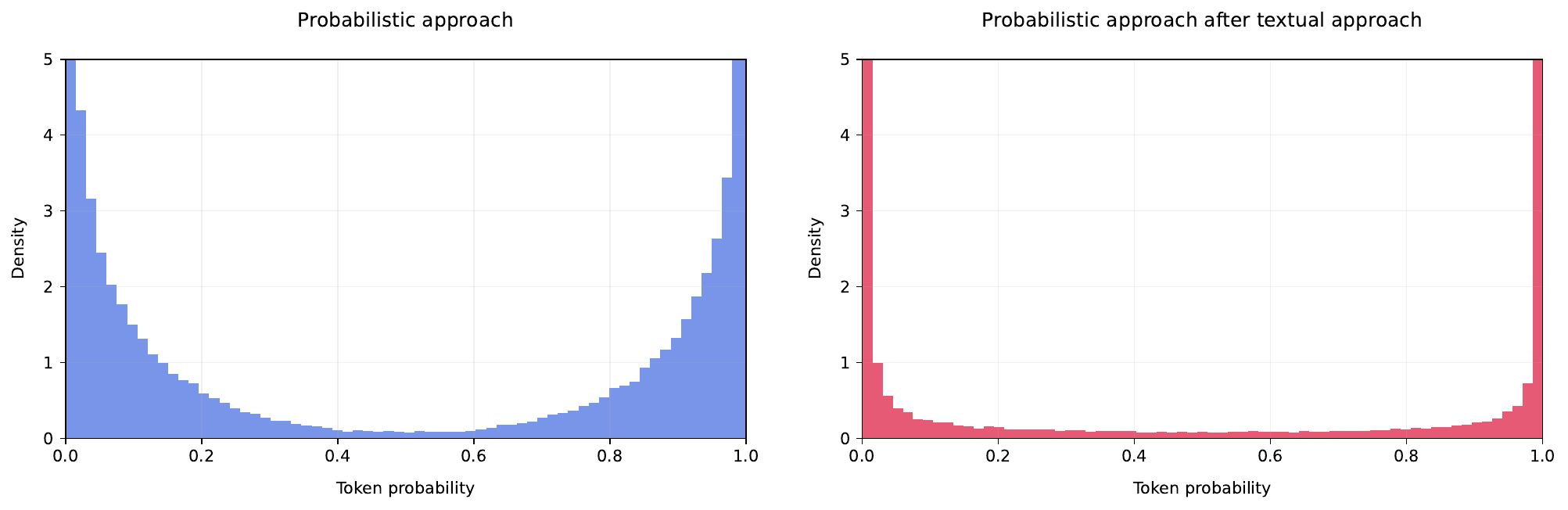}
    \caption{Distribution of response probabilities with and without reasoning}
    \label{fig:prob_after_reasoning}
\end{figure}
The initial distribution (Fig.~\ref{fig:prob_after_reasoning}) is multimodal in the form of $U$ with peaks at 0 and 1. LLM is generally confident in its answers and separates them well into two distinct buckets. When changing the approach, it can be seen that the examples in which the model is less confident become fewer because they diverge towards the peaks 0 and 1, and the answers become more deterministic. This is especially true for values in the range 1 $-$ $\epsilon$, where the threshold is selected, because the maximum probability among the other answers is taken, and the answers "yes" and "no" are opposite in meaning, and LLM tends to give the maximum probability to one or the another answer. Finally, the change in distribution means that the model is highly confident in its answer, since all the necessary reasoning for making a decision was provided, even if it was incorrect (model hallucinations); therefore, using classification thresholds for a probabilistic approach is no longer effective.

\begin{table}[H]
  \centering
  \footnotesize
  \caption{Metrics for prob-approach after text-approach (Binary)}
  \label{tab:prob_approach_after_text}
  \begin{tabularx}{\textwidth}{@{}l *{7}{c}@{}}
    \toprule
    \textbf{Model} &
    \makecell{\textbf{Prec.}\\\textit{1, pos.}} &
    \makecell{\textbf{Prec.}\\\textit{0, neg.}} &
    \makecell{\textbf{Rec.}\\\textit{1, pos.}} &
    \makecell{\textbf{Rec.}\\\textit{0, neg.}} &
    \textbf{Acc.} &
    \makecell{\textbf{f1-score}\\ (macro avg.)} &
    \makecell{\textbf{Coverage}\\ (perc.)} \\
    \midrule
    \makecell[l]{Qwen2.5-32b\\text-approach}                
    & 0.8613 & 0.6531 & \textbf{0.8726} & 0.6307 & 0.8059 &  0.7543 & 100\% \\
    \midrule
    \makecell[l]{Qwen2.5-32b \\prob-approach\\use thresholds}                   & \textbf{0.9398} & \textbf{0.6863} & 0.8518 & \textbf{0.8560 }& \textbf{ 0.8530} &  \textbf{0.8277} & 54\%\\
    \midrule
    \makecell[l]{Qwen2.5-32b \\text-app.+prob-app}                & 0.8507  & 0.6540 &  0.8678 & 0.6214 &  0.7971 & 0.7482 & 100\% \\
    \midrule
    \makecell[l]{Qwen2.5-32b \\text-app.+prob-app.\\use thresholds}                & 0.8614 & 0.6628 & 0.7285 & 0.8199 & 0.7646 & 0.7612 & 47\% \\
    \bottomrule
  \end{tabularx}
\end{table}
Table~\ref{tab:prob_approach_after_text} shows that the metrics of the approach using model reasoning and answer probability thresholds do not significantly improve the model metrics (metrics are still close to the text-approach and lower than the classic probability approach), even though 53\% of the less confident predictions were marked as \textit{"unk"}. A probabilistic approach without reasoning provides better metrics.

\subsection{RAG}
\subsubsection{Multi-class}
\begin{table}[H]
  \centering
  \footnotesize
  \caption{Metrics for RAG (Multi-class, text-approach)}
  \label{tab:rag_multiclass}
  \begin{tabularx}{\textwidth}{@{}l *{5}{c}@{}}
    \toprule
    \textbf{Model} &
    \makecell{\textbf{Coverage}\\(\textit{conf}=\textit{"unk"})} &
    \makecell{\textbf{Coverage}\\(\textit{conf}\,\(\neq\)\,\textit{"unk"})} &
    \makecell{\textbf{Accuracy}\\(\textit{w/o conf})} &
    \makecell{\textbf{Accuracy}\\(\textit{conf}=\textit{wrong})} &
    \textbf{$\text{Accuracy}_{\mathrm{total}}$} \\
    \midrule
    \textbf{Human}    & 0.2134 & 0.5240 & \textbf{0.6977} & 0.3113 & 0.1489 \\    
    \midrule
    \makecell[l]{Qwen2.5-32b}                & 0.5243 & 0.5243 &  0.5399 &  \textbf{0.5399} & 0.2831 \\
    \midrule
    \makecell[l]{Llama3.3-70b}       & 0.6585 & 0.6585 & 0.4118 & 0.4118 & 0.2712 \\
    \midrule
    \makecell[l]{Llama3.3-70b\\\textit{retriever:}\\multilingual-e5-large-instruct\\Find 1 document}     & 0.6479 & 0.6479 & 0.4221 & 0.4221 & 0.2735 \\
    \midrule
    \makecell[l]{Llama3.3-70b\\\textit{retriever:}\\multilingual-e5-large-instruct\\Find 5 documents}
    & \textbf{0.7062} & \textbf{0.7062} & 0.4013 & 0.4013 &  0.2834 \\
    \midrule
    \makecell[l]{Llama3.3-70b\\\textit{retriever:}\\BM25\\Find 1 document}                & 0.6440 & 0.6440 & 0.4209 & 0.4209 & 0.2711 \\
    \midrule
    \makecell[l]{Llama3.3-70b\\\textit{retriever:}\\BM25\\Find 5 documents}                & 0.6234 & 0.6234 & 0.4333 & 0.4333 & 0.2701 \\
    \midrule
    \makecell[l]{Llama3.3-70b\\\textit{retriever:}\\multilingual-e5-large-instruct\\\textit{reranker:}\\frida\\Find 5 from 25 documents}     & 0.6416 & 0.6416 & 0.4517 & 0.4517 & 0.2898 \\
    \midrule
    \makecell[l]{Llama3.3-70b\\\textit{retriever:}\\multilingual-e5-large-instruct\\\textit{reranker:}\\bge-m3\\Find 5 from 25 documents\\\textbf{best RAG}}    & 0.6446 & 0.6446 & 0.4519 & 0.4519 & 0.2913 \\
    \midrule
    \makecell[l]{Qwen2.5-32b\\\textit{retriever:}\\multilingual-e5-large-instruct\\\textit{reranker:}\\bge-m3\\Find 5 from 25 documents\\\textbf{best RAG}}     & 0.5483 & 0.5483 &  0.5334 &  0.5334 &\textbf{ 0.2925} \\
    \bottomrule
  \end{tabularx}
\end{table}

Table~\ref{tab:rag_multiclass} shows that for Llama3.3–70b with the "multilingual-e5-large-instruct" retriever, retrieving 5 documents slightly boosts coverage compared to 1 document, but accuracy dips. The net effect is a negligible \textbf{$\text{Accuracy}_{\mathrm{total}}$} change from 0.2735 to 0.2834.
A reverse pattern holds for BM25 retrieval: more docs slightly improve accuracy despite a drop in coverage, but \textbf{$\text{Accuracy}_{\mathrm{total}}$} remains roughly flat. The best quality is achieved when using a retriever and a re-ranker together ("multilingual-e5-large-instruct" and "bge-m3"). A good reranker together with a retriever in both cases under consideration resulted in an increase in accuracy or coverage metrics. The maximum increase in \textbf{$\text{Accuracy}_{\mathrm{total}}$} are seen with Llama3.3-70b from 0.2712 to 0.2925.

\begin{table}[H]
  \centering
  \footnotesize
  \caption{Metrics for RAG few-shot for intents (Multi-class, text-approach)}
  \label{tab:rag_multiclass_intent}
  \begin{tabularx}{\textwidth}{@{}l *{5}{c}@{}}
    \toprule
    \textbf{Model} &
    \makecell{\textbf{Coverage}\\(\textit{conf}=\textit{"unk"})} &
    \makecell{\textbf{Coverage}\\(\textit{conf}\,\(\neq\)\,\textit{"unk"})} &
    \makecell{\textbf{Accuracy}\\(\textit{w/o conf})} &
    \makecell{\textbf{Accuracy}\\(\textit{conf}=\textit{wrong})} &
    \textbf{$\text{Accuracy}_{\mathrm{total}}$} \\
    \midrule
    \textbf{Human}    & 0.2134 & 0.5240 & \textbf{0.6977} & 0.3113 & 0.1489 \\    
    \midrule
    \makecell[l]{Llama3.3-70b}       & 0.6585 & 0.6585 & 0.4118 & \textbf{0.4118} &\textbf{ 0.2712} \\
    \midrule
    \makecell[l]{Llama3.3-70b\\\textit{retriever:}\\multilingual-e5-large-instruct\\Find 1 documents}     & \textbf{0.6645} & \textbf{0.6645} & 0.3669 & 0.3669 & 0.2438 \\
    \midrule
    \makecell[l]{Llama3.3-70b\\\textit{retriever:}\\ multilingual-e5-large-instruct\\Find 5 documents}     & 0.6427 & 0.6427 & 0.3141 & 0.3141 & 0.2018 \\
    \bottomrule
  \end{tabularx}
\end{table}

Table~\ref{tab:rag_multiclass_intent} shows that searching for a document for each intent decreases \textbf{$\text{Accuracy}_{\mathrm{total}}$}. This is most likely because either 1 or 5 documents are added for each intent, which greatly expands the prompt and risks LLM hallucinations. The analysis also showed that some documents may contain inconsistent information. This is more noticeable when a large number of documents are placed in the prompt than when searching for relevant documents for a single client request.

\subsubsection{Binary}
\begin{table}[H]
  \centering
  \footnotesize
  \caption{Metrics for RAG (Binary)}
  \label{tab:rag_binary}
  \begin{tabularx}{\textwidth}{@{}l *{7}{c}@{}}
    \toprule
    \textbf{Model} &
    \makecell{\textbf{Prec.}\\\textit{1, pos.}} &
    \makecell{\textbf{Prec.}\\\textit{0, neg.}} &
    \makecell{\textbf{Rec.}\\\textit{1, pos.}} &
    \makecell{\textbf{Rec.}\\\textit{0, neg.}} &
    \textbf{Acc.} &
    \makecell{\textbf{f1-score}\\ (macro avg.)} &
    \makecell{\textbf{Coverage}\\ (perc.)} \\
    \midrule
    \textbf{Human}    & 0.8748 & 0.7162 & 0.7834 & 0.8297 & 0.8018 & 0.7977 & 100\% \\
     \midrule
    \makecell[l]{Llama3.3-70b\\text-approach}          &0.8779 & 0.8178 & \textbf{0.9743} & 0.4600 & 0.8711 & 0.7562&100\%\\
     \midrule
     \makecell[l]{Qwen2.5-32b\\text-approach}        & 0.8613 & 0.6531 & 0.8726 & 0.6307 & 0.8059 &  0.7543 & 100\% \\
     \midrule 
    \makecell[l]{Llama3.3-70b\\prob-approach\\use thresholds}                & 0.8918 & 0.6825 & 0.7387 & 0.8624 & 0.7875 & 0.7850 & 57\% \\
    \midrule
    \makecell[l]{Qwen2.5-32b \\prob-approach\\use thresholds}                   & \textbf{0.9398} & 0.6863 & 0.8518 & 0.8560 &  0.8530 &  \textbf{0.8277} & 54\%\\
\midrule 
    \makecell[l]{Llama3.3-70b\\\textit{retriever:}\\multilingual-e5-large-instruct\\\textit{reranker:}\\bge-m3\\Find 5 from 25 documents\\\textbf{best RAG}\\prob-approach\\use thresholds}  & 0.8835 & 0.7416 & 0.7674 & \textbf{0.8683} & 0.8113 & 0.8106 & 61\%\\ 
     \midrule
  \makecell[l]{Qwen2.5-32b\\\textit{retriever:}\\multilingual-e5-large-instruct\\\textit{reranker:}\\bge-m3\\Find 5 from 25 documents\\\textbf{best RAG}\\text-approach}     & 0.8787 & 0.6932 & 0.7591 & 0.8386 & 0.7904 & 0.7868 & 100\% \\
  \midrule
    \makecell[l]{Qwen2.5-32b\\\textit{retriever:}\\multilingual-e5-large-instruct\\\textit{reranker:}\\bge-m3\\Find 5 from 25 documents\\\textbf{best RAG}\\prob-approach\\use thresholds}                & 0.8993 &  \textbf{0.8401} & 0.9712 & 0.5818 &\textbf{ 0.8908} & 0.8103 & 59\% \\
    
    \bottomrule
  \end{tabularx}
\end{table}
In the case of binary annotation (Table~\ref{tab:rag_binary}), adding the best RAG approach from multi-class annotating increases the annotation metrics for Llama3.3-70b prob-approach except for a slight drop in precision on positives (probably due to the strong increase in precision on negatives). Qwen2.5-32b also shows a significant increase in annotation metrics, but a severe drop in recall on negatives. We should also notice an increase in coverage for both LLMs, especially Qwen2.5-32b, which shows that adding more information increases the LLM's confidence in the answer while not decreasing the other metrics.

\subsection{SFT (classification)}

\begin{table}[H]
  \centering
  \footnotesize
  \caption{Metrics after SFT (classification) (Multi-class)}
  \label{tab:sft_classification}
  \begin{tabularx}{\textwidth}{@{}l *{5}{c}@{}}
    \toprule
    \textbf{Model} &
    \makecell{\textbf{Coverage}\\(\textit{conf}=\textit{"unk"})} &
    \makecell{\textbf{Coverage}\\(\textit{conf}\,\(\neq\)\,\textit{"unk"})} &
    \makecell{\textbf{Accuracy}\\(\textit{w/o conf})} &
    \makecell{\textbf{Accuracy}\\(\textit{conf}=\textit{wrong})} &
    \textbf{$\text{Accuracy}_{\mathrm{total}}$} \\
    \midrule
    \textbf{Human}                     & 0.2134 & 0.5240 & \textbf{0.6977} & 0.3113 & 0.1489  \\
    \midrule
    \makecell[l]{Llama3.3-70b\\text-approach}      
     & 0.6585 & 0.6585 & 0.4118 & 0.4118 & 0.2712 \\
     \midrule
    \makecell[l]{Llama3.3-70b\\text-approach\\best RAG}               & 0.6446 & 0.6446 & 0.4519 & \textbf{0.4519} & \textbf{0.2913} \\
    \midrule
     \makecell[l]{Llama3.3-70b\\LoRA\\r=4, alpha=8\\(1 epoch)}     & 0.6712 & 0.6712 & 0.4055 & 0.4055 & 0.2722 \\
    \midrule
\makecell[l]{Llama3.3-70b\\LoRA\\r=4, alpha=8\\(2 epochs)}     & \textbf{0.7328} & \textbf{0.7328} & 0.3455 & 0.3455 & 0.2532\\
    \bottomrule
  \end{tabularx}
\end{table}
As a result of training (Table~\ref{tab:sft_classification}), the model began to choose unk a more rarely, which led to an increase in coverage but a decrease in accuracy on all other intents. It was also found that the probability distribution had shifted in the direction described in Section~\ref{sec:prob_after_text}, which also made the thresholds more difficult to choose. RAG provides better total accuracy than this type of adaptation. Based on the results obtained on the multi-class data and the impossibility of obtaining explanations from the model using this approach, experiments on binary annotation were conducted using the method described in Section~\ref{sec:sft_reasoning_method}.

\subsection{SFT (reasonings)}\label{sec:sft_reasonings}

\begin{table}[H]
  \centering
  \footnotesize
  \caption{Metrics after SFT (reasonings) (Binary, Prob-approach)}
  \label{tab:sft_reasonings}
  \begin{tabularx}{\textwidth}{@{}l *{7}{c}@{}}
    \toprule
    \textbf{Model} &
    \makecell{\textbf{Prec.}\\\textit{1, pos.}} &
    \makecell{\textbf{Prec.}\\\textit{0, neg.}} &
    \makecell{\textbf{Rec.}\\\textit{1, pos.}} &
    \makecell{\textbf{Rec.}\\\textit{0, neg.}} &
    \textbf{Acc.} &
    \makecell{\textbf{f1-score}\\ (macro avg.)} &
    \makecell{\textbf{Coverage}\\ (perc.)} \\
    \midrule
    \textbf{Human}    & 0.8748 & 0.7162 & 0.7834 & 0.8297 & 0.8018 & 0.7977 & 100\% \\
\midrule
     \makecell[l]{Qwen2.5-32b\\text-approach\\best RAG}     & 0.8787 & 0.6932 & 0.7591 & 0.8386 & 0.7904 & 0.7868 & 100\% \\
\midrule
    \makecell[l]{Qwen2.5-32b\\text-approach\\LoRA\\\textit{r=4, alpha=8}\\\textit{dropout=0.1}\\best RAG}                      &  0.8947 & 0.7058 & 0.7565 & 0.8677 & 0.8012 & 0.7991 & 100\% \\

\midrule
     \makecell[l]{Qwen2.5-32b\\prob-approach\\use thresholds\\best RAG}                   & 0.8993 &  \textbf{0.8401} & \textbf{0.9712} & 0.5818 & \textbf{0.8908} & 0.8103 & 59\% \\
\midrule
    \makecell[l]{Qwen2.5-32b\\prob-approach\\use thresholds\\LoRA\\\textit{r=4, alpha=8}\\\textit{dropout=0.1}\\best RAG\\\textbf{best LoRA}}                     & \textbf{0.9131} & 0.7641 & 0.8131 & \textbf{0.8867} & 0.8430 & \textbf{0.8405 }& 63\% \\
    
    \bottomrule
  \end{tabularx}
\end{table}

As shown in Table~\ref{tab:sft_reasonings}, the metrics improved not only on the text-based approach (which is the original training goal), but also on the prob-approach. Interesting how much the recall on negatives has increased in the probabilistic approach. If before for good recall on negatives reasoning from LLM in text-approach mode, now on probabilistic approach a comparable recall on negatives is obtained. High-quality reasoning has indeed improved LLM metrics in the annotation task. it is also important to remember that RAG documents were added to the prompt during retraining. The improvement in metrics can also be attributed to the correct handling of retrieved documents, which Qwen2.5-32b learned from Deepseek-r1 and GPT-4o.

\subsection{Final best approaches}
\subsubsection{Multi-class}

\begin{table}[H]
  \centering
  \footnotesize
  \caption{Metrics for ensemble (Multi-class, Prob-approach)}
  \label{tab:multiclass_ensemble}
  \begin{tabularx}{\textwidth}{@{}l *{5}{c}@{}}
    \toprule
    \textbf{Model} &
    \makecell{\textbf{Coverage}\\(\textit{conf}=\textit{"unk"})} &
    \makecell{\textbf{Coverage}\\(\textit{conf}\,\(\neq\)\,\textit{"unk"})} &
    \makecell{\textbf{Accuracy}\\(\textit{w/o conf})} &
    \makecell{\textbf{Accuracy}\\(\textit{conf}=\textit{wrong})} &
    \textbf{$\text{Accuracy}_{\mathrm{total}}$} \\
     \midrule
    \makecell[l]{Qwen2.5-32b best-RAG\\prob approach\\use thresholds}               & \textbf{0.4376} & \textbf{0.4376} &  0.6831 & 0.6831 & 0.2989 \\
     \midrule
    \makecell[l]{Llama3.3-70b best-RAG\\prob approach\\use thresholds}   & 0.4374 & 0.4374 & 0.6758 & 0.6758 & 0.2957 \\
    \midrule
    \textbf{Human}                     & 0.2134 & 0.524 & 0.6977 & 0.3113 & 0.1489  \\
    \midrule
    \makecell[l]{\textbf{Ensemble}\\Llama3.3-70b best-RAG\\\&\\Qwen2.5-32b best-RAG\\prob approach\\use thresholds}     & 0.4269 & 0.4269 &\textbf{ 0.7030} & \textbf{0.7030} & \textbf{0.3001} \\
    \bottomrule
  \end{tabularx}
\end{table}

The Table~\ref{tab:multiclass_ensemble} shows the best approach, which consists of averaging the prediction probabilities of two LLMs and applying a tuned threshold. By slightly reducing the coverage with threshold, it is possible to achieve an increase in the accuracy of the LLM separately and of the ensemble as a whole. We achieved approximately the same accuracy on intents as human annotators and the highest coverage. 

\subsubsection{Binary}
\begin{figure}[H]
    \centering
    \includegraphics[width=0.6\linewidth]{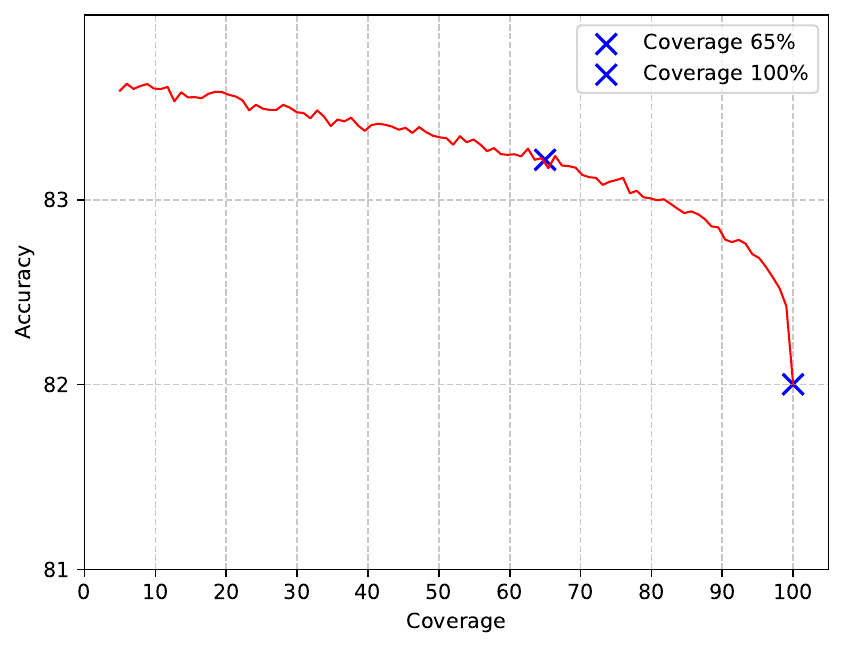}
    \caption{Example of accuracy dependence on coverage for a binary ensemble}
    \label{fig:thresholds}
\end{figure}

\begin{table}[H]
  \centering
  \footnotesize
  \caption{Metrics for ensemble (Binary, Prob-approach)}
  \label{tab:binary_ensemble}
  \begin{tabularx}{\textwidth}{@{}l *{7}{c}@{}}
    \toprule
    \textbf{Model} &
    \makecell{\textbf{Prec.}\\\textit{1, pos.}} &
    \makecell{\textbf{Prec.}\\\textit{0, neg.}} &
    \makecell{\textbf{Rec.}\\\textit{1, pos.}} &
    \makecell{\textbf{Rec.}\\\textit{0, neg.}} &
    \textbf{Acc.} &
    \makecell{\textbf{f1-score}\\ (macro avg.)} &
    \makecell{\textbf{Coverage}\\ (perc.)} \\
    \midrule
    \makecell[l]{Qwen2.5-32b\\best-RAG best-LoRA}            & \textbf{0.9131} & 0.7641 & 0.8131 &\textbf{ 0.8867} &\textbf{ 0.8430} &\textbf{ 0.8405} & 63\% \\
     \midrule
    \makecell[l]{Llama3.3-70b best-RAG}                & 0.8835 & 0.7416 & 0.7674 & 0.8683 & 0.8113 & 0.8106 & 61\%\\
\midrule
    \textbf{Human}    & 0.8748 & 0.7162 & 0.7834 & 0.8297 & 0.8018 & 0.7977 & 100\% \\
\midrule
    \makecell[l]{\textbf{Human}\\Based on the same\\ 65\% data as LLM}    & 0.8808 & 0.7403 &  0.8049 & 0.8362 & 0.8174 & 0.8133 & 65\% \\
    \midrule

    \makecell[l]{\textbf{Ensemble}\\Llama3.3-70b best-RAG\\\&\\Qwen2.5-32b\\best-RAG best-LoRA}                & 0.8851 &  \textbf{0.7680} &  \textbf{ 0.8271 }& 0.8421 & 0.8332 & 0.8292 & 65\% \\

    \bottomrule
  \end{tabularx}
\end{table}
In the case of binary annotation (Table~\ref{tab:binary_ensemble}), many of the approaches described in the paper outperformed human metrics; therefore, the goal was to increase coverage without significantly lowering metrics. To do this, two LLMs were also taken, whose probability predictions were averaged and a selected threshold was applied to them. The quality of annotation on the same data (65\%) as humans is higher for LLMs. This may mean that LLM annotates simple tasks better than humans, but with more complex tasks (where the probability of answers is below the threshold), human annotators still perform better. \\\indent
Fig.~\ref{fig:thresholds} demonstrates that the increase in accuracy with a decrease in the threshold (decrease in coverage) is non-linear.

\pagebreak
\section{Discussion}

Two approaches combining LLMs with RAG were proposed: binary pipeline uses two LLMs (Llama3.3-70b, Qwen2.5-32b), and one of these (Qwen2.5-32b) was fine‑tuned on the reasoning outputs of two large proprietary models. Considering annotation metrics (primarily accuracy and coverage), we replaced 65\% of binary annotations with LLMs and all multiclass annotations with LLMs, accelerating annotation speed by an average factor of six while improving overall quality.

The addition of RAG and thresholding affected each LLM differently; thus, there is no universal configuration, and applying these methods to new data or in another company will require repeated experimentation. Adapting an LLM purely as a classifier did not yield the same gains as fine‑tuning an LLM for reasoning, although the reasoning‑trained LLM also showed improved performance in the prob-approach—even when reasoning generation is disabled by design.

The results of reasoning fine‑tuning underscore the importance of correct initial prompt processing by the LLM and suggest that the model may "know" the correct answer from the very start of its generation.\\
\textbf{The focus of future work should be on:}
\begin{enumerate}
  \item \textbf{Domain‑specific SFT.}  
    Both LoRA fine‑tuning variants (for classification and for reasoning) were trained on our in‑house support‑chat data. Their benefits may not transfer to new domains or intent taxonomies without additional fine‑tuning, demanding per‑domain annotation efforts and compute resources.

  \item \textbf{Untrained RAG re‑ranker.}  
    Our RAG system employs a generic cross‑encoder (or BM25) for re‑ranking rather than one trained on our customer‑support corpus. A domain‑specific re‑ranker could reduce noise in retrieved documents and further improve annotation accuracy, especially for rare or nuanced intents.

  \item \textbf{Threshold sensitivity.}  
    Abstention thresholds for \textit{"unk"} were hand‑tuned for each LLM and benchmark. Because token‑probability calibration varies by model, prompt, and data distribution, these thresholds are unlikely to generalize across languages, models, or new datasets without ongoing monitoring and re‑tuning.

\item \textbf{No end-to-end measurements.} For optimal demonstration of the method's effectiveness, an experiment comparing two models trained only on human annotations or only on LLM annotations would be required, with the comparison conducted on a high-quality benchmark by highly paid professional annotators.

\item \textbf{Full distillation of the LLM.} There may be a way to explore knowledge-distillation or structured pruning to compress the LLM while retaining cross-domain annotation accuracy.

\item \textbf{RAG evaluation.} In Section~\ref{sec:sft_reasonings} reported metric gains. It will be interesting to look at specialized metrics for RAG systems. It was not only the correct reasoning that could have influenced the improvements, but specifically the ability to work correctly with the information received from RAG.
\end{enumerate}

These observations suggest that, while LLM‑based annotation can match or exceed human performance in our setting, deploying similar pipelines elsewhere will require careful domain adaptation, re‑ranking training, threshold calibration, and ongoing prompt and model optimization.
\pagebreak
\section{Conclusion}

In this work, we have demonstrated that human annotators in a large‑scale, multi‑intent text‑classification pipeline can be effectively replaced (or substantially supplemented) by large language models (LLMs) without sacrificing annotation quality. 

Starting from a labor‑intensive, expert‑driven benchmark of roughly 250 distinct client intents, we developed and compared two LLM‑based annotation paradigms:
\begin{itemize}
  \item A \emph{text‑approach}, in which the LLM generates natural‑language reasoning and a label.
  \item A \emph{prob‑approach}, in which the LLM emits exactly one token (or digit), and we interpret its token probability to decide whether to accept or defer to a human.
\end{itemize}
We further augmented these methods with:  
\begin{itemize}
  \item Soft fine‑tuning via LoRA for both classification heads and reasoning generation.  
  \item Retrieval‑augmented generation (RAG) to surface up‑to‑date. domain‑specific documents, and  
  \item Simple LLM ensembles to boost annotation confidence.  
\end{itemize}

Across binary and multi‑class benchmarks, our best LLM pipelines achieved annotation accuracy at least equal to—and in many cases exceeding—regular human annotators when compared against expensive expert benchmarks. Moreover, LLM annotation significantly reduced the average annotation time.

These findings suggest that LLMs can not only validate classifier predictions but also serve as primary annotators in production systems, enabling continuous, high‑quality incremental learning at scale. Future work will explore adaptive thresholding strategies, domain transfer to new intent taxonomies, and tighter integration of LLM‑based annotators within active‑learning loops to further improve both coverage and accuracy.

\newpage 

\printbibliography[heading=bibintoc]   % [] are for appearing in Table of Contents

\newpage
\appendix

\section{Appendix}\label{sec:appendix_prompt}
\begin{center}
\begin{minipage}{1\textwidth}
\begin{framed}
\noindent\textbf{Multi-class prompt} \\
\footnotesize
\medskip % Small vertical space

Hello! Here are the instructions you should follow when communicating with me:\\
1. You are an intelligent data annotation assistant. You receive text that a user has written in the app's support chat, along with several hypothetical intents (classes) that this text may refer to. Along with the intent names, you will also receive a human-written description of the intent and typical examples, also selected by a human. Carefully read and compare the intent description and examples with the text that needs to be classified.\\
2. it is very important that you follow the instructions carefully and are very careful when making decisions. \\
3. For each correctly completed task, you will receive bonuses that you can spend on self-development and training.\\
4. If you are able to classify the text clearly, enter the correct class after the [ANSWER] tag. Use the exact class name that was provided in the list, as this is necessary for subsequent quality assessment.\\
5. The text may not belong to any of the provided classes or may not contain any intent at all. In this case, display the answer in the format [ANSWER] \textit{"unk"} (\textit{"unk"} number from the task list).\\
6. If the context of the user's message is insufficient to provide an unambiguous answer using the provided list of classes, display the answer in the format [ANSWER] \textit{"unk"} (\textit{"unk"} number from the task list). Please note that this intent should only be used in extreme cases when you find it difficult to choose between several intents. This is a rare situation; therefore, read the intent descriptions carefully and match them to the client's phrase. \\
7. Write and reason in Russian. If you find this difficult, do it as you see fit and then translate your thoughts into Russian. \\
8. When solving tasks, be sure to write down all your reasoning in the text and analyze it while writing your answer.\\
9. Stick to the required output format:\\
1. [TEXT] user's text
2. [REASONING] Reasoning regarding the selection of the appropriate class. Reason consistently, consider all possible options, and use the class description and examples provided for it.\\
3. [ANSWER] Answer - the number of one of the intents in the list. If you decide that the answer should be one of the intents in the list, make sure you answer **ONLY** with the number of that intent.\\
10. You may also encounter additional data enclosed in the [RETREIVED] tag. This data was obtained from the internal information system and may contain additional useful information about the user's text.
\medskip

\end{framed}
\end{minipage}
\end{center}

\pagebreak
\begin{center}
\begin{minipage}{1\textwidth}
\begin{framed}
\noindent\textbf{Multi-class prob-approach prompt} \\
\footnotesize
\medskip % Small vertical space

Hello! Here are the instructions you should follow when communicating with me:\\
1. You are an intelligent data annotation assistant. You receive text that a user has written in the app's support chat, along with several hypothetical intents (classes) that this text may refer to. Along with the intent names, you will also receive a human-written description of the intent and typical examples, also selected by a human. Carefully read and compare the intent description and examples with the text that needs to be classified.\\
2. it is very important that you follow the instructions carefully and are very careful when making decisions. \\
3. For each correctly completed task, you will receive bonuses that you can spend on self-development and training.\\
4. If you are able to classify the text clearly, enter the correct class after the [ANSWER] tag. Use the exact number that was provided in the list, as this is necessary for subsequent quality assessment.\\
5. The text may not belong to any of the provided classes or may not contain any intent at all. In this case, enter your answer in the format [ANSWER] \textit{"unk"} (\textit{"unk"} number from the task list).\\
6. If the context of the user's message from [TEXT] is insufficient to provide a clear answer using the provided list of classes, enter your answer in the format [ANSWER]: \textit{"unk"} (\textit{"unk"} number from the task list). Please note that this intent should only be used in extreme cases when you find it difficult to choose between several intents. This is a rare situation; therefore, read the intent descriptions carefully and match them to the user's phrase.\\
7. Stick to the required output format: [ANSWER] Answer - number of one of the intents in the list. If you decide that the answer should be one of the intents in the list, make sure you respond with **ONLY** the number of that intent.\\
8. You may also encounter additional data enclosed in the [RETREIVED] tag. This data was obtained from the internal information system and may contain additional useful information about the user's text.
\medskip

\end{framed}
\end{minipage}
\end{center}

\pagebreak
\begin{center}
\begin{minipage}{1\textwidth}
\begin{framed}
\noindent\textbf{Binary prompt} \\
\footnotesize
\medskip % Small vertical space

Hello! Here are the instructions you should follow when communicating with me: \\
1. You are an intelligent data annotation assistant. You receive text that a user has written in the application's support chat and the suggested class to which this text may belong. You need to determine whether the text actually belongs to the suggested class or not. In addition to the class descriptions written by a human, you will also receive typical examples, also selected by a human. Read carefully and compare the class descriptions and examples with the text that needs to be classified. \\
2. it is very important that you follow the instructions carefully and are very careful when making decisions. \\
3. For each correctly completed task, you will receive bonuses that you can spend on self-development and training.\\
4. Write and think in Russian. If you find this difficult, do it as you see fit and then translate your thoughts into Russian.\\
5. Please note that the task text contains both suitable and unsuitable examples. Carefully compare the text with the examples to avoid falling into a trap. it is also important to consider product names or additional external context. If a class refers to one product or situation, but the text mentions another product or situation, this means that the text does not fit the specified class. \\
6. When solving problems, be sure to write down all your reasoning in the text and analyze it while writing your answer. Reason step by step; for convenience, you can break your reasoning down into separate points. \\
7. Follow the required format for presenting your conclusions: \\
1. [TEXT] User's text \\
1. [REASONING] Reasoning used to verify the correctness of the prediction. Reason consistently, step by step, consider all possible options, and be sure to use the class description and examples provided for it. Also, clarify the meaning of ambiguous words that may imply that the text does not belong to this class. Do not guess what the client means or try to find hidden meanings in their message. Evaluate whether a message belongs to a class based *only* on the text of the message. \\
2. [ANSWER] The answer is a single line "yes" or "no" in lowercase (small letters). If the text belongs to the intended class or is similar to the examples provided, answer "yes." If the text cannot be related to this class and is not similar to any of the examples given, answer "no." Be sure to give your final answer based on your previous reasoning. \\
8. You may also encounter additional data enclosed in the [RETREIVED] tag. This data was obtained from the internal information system and may contain additional useful information about the user's text.

\medskip

\end{framed}
\end{minipage}
\end{center}

\pagebreak
\begin{center}
\begin{minipage}{1\textwidth}
\begin{framed}
\noindent\textbf{Binary prob-approach prompt} \\
\footnotesize
\medskip % Small vertical space

Hello! Here are the instructions you should follow when communicating with me: \\
1. You are an intelligent data annotation assistant. You receive text that a user has written in the application's support chat and the suggested class to which this text may belong. You need to determine whether the text actually belongs to the suggested class or not. In addition to the class descriptions written by a human, you will also receive typical examples, also selected by a human. Read carefully and compare the class descriptions and examples with the text that needs to be classified. \\
2. it is very important that you follow the instructions carefully and are very careful when making decisions. \\
3. For each correctly completed task, you will receive bonuses that you can spend on self-development and training.\\
4. Please note that the task text contains both suitable and unsuitable examples. Carefully compare the text with the examples to avoid falling into a trap. it is also important to consider product names or additional external context. If a class refers to one product or situation, but the text mentions another product or situation, this means that the text does not fit the specified class. \\
5. Follow the required format for presenting your conclusions: 1. [TEXT] User's text, [ANSWER] The answer is a single line "0" or "1" in lowercase (small letters). If the text belongs to the intended class or is similar to the examples provided, answer "1." If the text cannot be related to this class and is not similar to any of the examples given, answer "0." \\
6. You may also encounter additional data enclosed in the [RETREIVED] tag. This data was obtained from the internal information system and may contain additional useful information about the user's text.

\medskip

\end{framed}
\end{minipage}
\end{center}
\end{document}